# Approximate Linear Programming for First-order MDPs

**Scott Sanner**
University of Toronto
Department of Computer Science
Toronto, ON, M5S 3H5, CANADA
ssanner@cs.toronto.edu

**Craig Boutilier**
University of Toronto
Department of Computer Science
Toronto, ON, M5S 3H5, CANADA
cebly@cs.toronto.edu

## Abstract

We introduce a new approximate solution technique for first-order Markov decision processes (FOMDPs). Representing the value function linearly w.r.t. a set of first-order basis functions, we compute suitable weights by casting the corresponding optimization as a *first-order linear program* and show how off-the-shelf theorem prover and LP software can be effectively used. This technique allows one to solve FOMDPs independent of a specific domain instantiation; furthermore, it allows one to determine bounds on approximation error that apply *equally to all domain instantiations*. We apply this solution technique to the task of elevator scheduling with a rich feature space and multi-criteria additive reward, and demonstrate that it outperforms a number of intuitive, heuristically-guided policies.

## 1 Introduction

Markov decision processes (MDPs) have become the *de facto* standard model for decision-theoretic planning problems. While classic dynamic programming algorithms for MDPs [14] require explicit state and action enumeration, recent techniques for exploiting propositional structure in factored MDPs [2] avoid explicit state and action enumeration, thus making it possible to optimally solve extremely complex problems containing $10^8$ states [9]. Approximation techniques for factored MDPs using basis functions have also been successful, yielding approximate solutions to problems with up to $10^{40}$ states [8, 17].

While such techniques for factored MDPs have proven effective, they cannot generally exploit first-order structure. Yet many realistic planning domains are best represented in first-order terms, exploiting the existence of domain objects, relations over these objects, and the ability to express objectives and action effects using quantification. As a result, a new class of algorithms have been introduced to explicitly handle MDPs with relational (RMDP) and first-order (FOMDP) structure.[1] *Symbolic dynamic programming (SDP)* [3], *first-order value iteration (FOVIA)* [10], and the *relational Bellman algorithm (ReBel)* [11] are model-based algorithms for solving FOMDPs and RMDPs, using appropriate generalizations of value iteration [14]. *Approximate policy iteration* [5] induces rule-based policies from sampled experience in small-domain instantiations of RMDPs and generalizes these policies to larger domains. In a similar vein, *inductive policy sslection using first-order regression* [6] uses regression to provide the hypothesis space over which a policy is induced. *Approximate linear programming (for RMDPs)* [7] is an approximation technique using linear program optimization to find a best-fit value function over a number of sampled RMDP domain instantiations.[2]

In this paper, we provide a novel approach to approximation of FOMDP value functions by representing them as a linear combination of first-order basis functions and using a first-order generalization of approximate linear programming techniques for propositional MDPs [7, 17] to solve for the required weights. While many FOMDP solution techniques need to instantiate a ground FOMDP w.r.t. a set of domain objects to obtain a propositional version of the FOMDP (usually performing this grounding operation multiple times for many different domain sizes), our approach exploits the power of full first-order reasoning taken by SDP, thus avoiding the exponential blowup that can result from domain instantiation. Since the use of first-order reasoning techniques allows our solution to abstract over all possible ground domain instantiations, the bounds that we derive for approximation error apply *equally to all domain instantiations* (i.e., the bounds are independent of domain size and and thus do not rely on a distribution over domain sizes).

---

[1] We use the term *relational MDP* to refer models that allow implicit existential quantification, and *first-order MDP* for those with explicit existential and universal quantification.

[2] Because the language in this work allows count aggregators in the transition and reward functions, it is more expressive than typical RMDPs.

## 2 Markov Decision Processes

Formally, a finite state and action MDP [14] is a tuple $\langle S, A, T, R \rangle$ where: $S$ is a finite state space; $A$ is a finite set of actions; $T : S \times A \times S \rightarrow [0, 1]$ is a transition function, with $T(s, a, \cdot)$ a distribution over $S$ for all $s \in S, a \in A$; and $R : S \times A \rightarrow \mathbb{R}$ is a bounded reward function. Our goal is to find a policy $\pi$ that maximizes the value function, defined using the infinite horizon, discounted reward criterion: $V_\pi(s) = E_\pi[\sum_{t=0}^{\infty} \gamma^t \cdot r^t | S_0 = s]$, where $r^t$ is a reward obtained at time $t$, $0 \leq \gamma < 1$ is a discount factor, and $S_0$ is the initial state.

A stationary policy takes the form $\pi : S \rightarrow A$, with $\pi(s)$ denoting the action to be executed in state $s$. Policy $\pi$ is optimal if $V_\pi(s) \geq V_{\pi'}(s)$ for all $s \in S$ and all policies $\pi'$. The optimal value function $V^*$ is the value of any optimal policy and satisfies the following:

$$V^*(s) = R(s, a) + \max_a \gamma \sum_{t \in S} T(s, a, t) \cdot V^*(t) \quad (1)$$

In the approximate linear programming solution to MDPs [8, 17] on which our work is based, we represent the value function as the weighted sum of $k$ basis functions $B_i(s)$:

$$V(s) = \sum_{i=1}^{k} w_i B_i(s) \quad (2)$$

Our goal is to find a good setting of weights that approximates the optimal value function. One way of doing this is to cast the optimization problem as a linear program that directly solves for the weights of an $L_1$-minimizing approximation of the optimal value function:

Variables: $w_1, \ldots, w_k$

Minimize: $\sum_{s \in S} \sum_{i=1}^{k} w_i B_i(s)$

Subject to: $0 \geq \left( R(s, a) + \gamma \sum_{t \in S} T(s, a, t) \cdot \sum_{i=1}^{k} w_i B_i(t) \right) - \sum_{i=1}^{k} w_i B_i(s) \; ; \; \forall a \in A, s \in S \quad (3)$

While the size of the objective and the number of constraints in this LP is proportional to the number of states (and therefore exponential), recent techniques use compact, factored basis functions and exploit the reward and transition structure of factored MDPs [8, 17], making it possible to avoid generating an exponential number of constraints (and rendering the entire LP compact).

## 3 First-Order Representation of MDPs

### 3.1 The Situation Calculus

The situation calculus [13] is a first-order language for axiomatizing dynamic worlds. Its basic ingredients consist of *actions*, *situations*, and *fluents*. Actions are first-order terms involving action function symbols. For example, the action of moving an elevator $e$ up one floor might be denoted by the action term *up(e)*. A situation is a first-order term denoting the occurrence of a sequence of actions. These are represented using a binary function symbol *do*: $do(\alpha, s)$ denotes the situation resulting from doing action $\alpha$ in situation $s$. In an elevator domain, the situation term

$$do(up(E), do(open(E), do(down(E), S_0)))$$

denotes the situation resulting from executing sequence *[down(E),open(E),up(E)]* in $S_0$. Relations whose truth values vary from state to state are called fluents, and are denoted by predicate symbols whose last argument is a situation term. For example, *EAt(e, 10, s)* is a relational fluent meaning that elevator $e$ is at floor 10 in situation $s$.[3]

A domain theory is axiomatized in the situation calculus with four classes of axioms [15]. For this paper, the most important of these axioms are the *successor state axioms (SSAs)*. There is one such axiom for each fluent $F(\vec{x}, s)$, with syntactic form $F(\vec{x}, do(a, s)) \equiv \Phi_F(\vec{x}, a, s)$ where $\Phi_F(\vec{x}, a, s)$ is a formula with free variables among $a, s, \vec{x}$. These characterize the truth values of the fluent $F$ in the next situation $do(a, s)$ in terms of the current situation $s$, and embody a solution to the frame problem for deterministic actions [15].

Following is a a simple example of an SSA for *EAt* (assuming 1 is the ground floor):

*EAt*$(e, 1, do(a, s)) \equiv$
 *EAt*$(e, 2, s) \land a = down(e) \lor$ *EAt*$(e, 1, s) \land a \neq up(e)$

This states that an elevator $e$ is at the first floor after doing action $a$ in situation $s$ if $e$ was at the second floor in $s$ and the action *down(e)*, or if $e$ was already at the first floor in $s$ and the action was not to move up.

The *regression* of a formula $\psi$ through an action $a$ is a formula $\psi'$ that holds prior to $a$ being performed iff $\psi$ holds after $a$. Successor state axioms support regression in a natural way. Suppose that fluent $F$'s SSA is $F(\vec{x}, do(a, s)) \equiv \Phi_F(\vec{x}, a, s)$. We inductively define the regression of a formula whose situation arguments all have the form $do(a, s)$ as follows:

$$\begin{aligned} Regr(F(\vec{x}, do(a, s))) &= \Phi_F(\vec{x}, a, s) \\ Regr(\neg \psi) &= \neg Regr(\psi) \\ Regr(\psi_1 \land \psi_2) &= Regr(\psi_1) \land Regr(\psi_2) \\ Regr((\exists x)\psi) &= (\exists x)Regr(\psi) \end{aligned}$$

---
[3]In contrast to states, situations reflect the entire history of action occurrences. As we will see, the specification of dynamics in the situation calculus will allow us to recover state properties from situation terms, and indeed, the form we adopt for successor state axioms renders the process Markovian.

For example, regressing $\neg EAt(e, 1, do(a, s))$ exploits the the SSA for $EAt$ to obtain the following description of the conditions on the "pre-action" situation $s$ under which this formula is true:

$Regr(\exists e \neg EAt(e, 1, do(a, s))) =$
$\exists e \neg[EAt(e, 2, s) \land a = down(e) \lor EAt(e, 1, s) \land a \neq up(e)]$

### 3.2 Stochastic Actions and the Situation Calculus

The classical situation calculus relies on the deterministic form of actions to allow compact and natural representation of dynamics using SSAs as described in the previous section. To generalize this to stochastic actions, as required by FOMDPs, we decompose stochastic actions (those in the control of some agent) into a *collection* of deterministic actions, each corresponding to a possible outcome of the stochastic action. We then specify a distribution according to which "nature" may choose a deterministic action from this set whenever that stochastic action is executed. As a consequence we need formulate SSAs using only these deterministic *nature's choices* [1, 3], obviating the need for a special treatment of stochastic actions in SSAs.

We illustrate this approach with a simple example in an elevator domain consisting of elevators, floors, and people. At each time step, an elevator can move up, down, or can open its doors at the current floor. Suppose the *open* action is stochastic with two possible outcomes, success and failure (we suppose *up* and *down* action are deterministic and always succeed). We decompose *open* into two deterministic choices, *openS* and *openF*, corresponding to success and failure. The probability of each choice is then specified, for example:

$EAt(e, 1, s) \supset prob(openS(e), open(e), s) = 0.85$
$EAt(e, 1, s) \supset prob(openF(e), open(e), s) = 0.15$
$EAt(e, f, s) \land f \neq 1 \supset prob(openS(e), open(e), s) = 0.9$
$EAt(e, f, s) \land f \neq 1 \supset prob(openF(e), open(e), s) = 0.1$
$prob(upS(e), up(e), s) = 1.0$
$prob(downS(e), down(e), s) = 1.0$

Here we see that the probability of the door failing to open is slightly higher on the first floor (0.15) than on other floors (0.10); this is simply to illustrate the "style" of the representation. Next, we define the SSAs in terms of these deterministic choices. However, in order to formalize the Elevator domain independent of any fixed set of floors, we must first introduce an axiomatization of above and below functions ($fa(x)$ and $fb(x)$ respectively) as well as transitive closure predicates $above(x, y)$ and $below(x, y)$ over these functions:[4]

[4]While transitive closure is technically not first-order definable, the axiomatization of $above(x, y)$ and $below(x, y)$ given here suffices to yield a contradiction when an assertion conflicts with an inference derivable from these axioms. Inconsistency detection is all that is needed by our algorithms.

$\forall x, y (\exists z\ below(x, z) \land below(z, y)) \supset below(x, y)$
$\forall x, y (\exists z\ above(x, z) \land above(z, y)) \supset above(x, y)$
$\forall x, y\ above(x, y) \supset \neg below(x, y)$
$\forall x, y\ above(x, y) \supset below(y, x)$
$\forall x, y\ below(x, y) \supset \neg above(x, y)$
$\forall x, y\ below(x, y) \supset above(y, x)$
$\forall x, y\ above(x, y) \supset x \neq y;\ \forall x, y\ below(x, y) \supset x \neq y$
$\forall x, y\ x = y \supset \neg above(x, y);\ \forall x, y\ x = y \supset \neg below(x, y)$
$\forall x\ above(fa(x), x);\ \forall x\ below(fb(x), x)$
$\forall x\ x = fa(fb(x));\ \forall x\ x = fb(fa(x))$

Given these foundational axioms, we can define the SSAs in a straightforward way.[5] For the fluent $EAt(e, f, s)$ (meaning elevator $e$ is at floor $f$), we have the following SSA:

$EAt(e, f, do(a, s)) \equiv$
$\quad EAt(e, fb(f), s) \land a = upS(e) \lor$
$\quad EAt(e, fa(f), s) \land a = downS(e) \lor$
$\quad EAt(e, f, s) \land a \neq upS(e) \land a \neq downS(e)$

Intuitively, $e$ is at floor $f$ after doing $a$ only if $e$ (successfully) moved up or down from an suitable floor above or below, or if $e$ was already at $f$ and did not (successfully) move up or down. The SSA $PAt(p, f, s)$ (person $p$ is at floor $f$) is:

$PAt(p, f, do(a, s)) \equiv$
$\quad [\exists e\ EAt(e, f, s) \land OnE(p, e, s) \land$
$\quad Dst(p, f) \land a = openS(e)] \lor [PAt(p, f, s) \land$
$\quad \neg(\exists e\ EAt(e, f, s) \land \neg Dst(p, f) \land a = openS(e))]$

Intuitively, person $p$ is at floor $f$ in the situation resulting from doing $a$ iff $p$ has destination $f$ and was on $e$, which had (stopped and successfully) opened its doors at $f$; or $p$ was at $f$ already and it's not the case that $e$ opened its doors at $f$ and $p$ was not at their destination. Finally, $OnE(p, e, s)$ (person $p$ is on elevator $e$) has the following SSA (with obvious meaning):

$OnE(p, e, do(a, s))$
$\quad [\exists f\ EAt(e, f, s) \land PAt(p, f, s) \land$
$\quad \neg Dst(p, f) \land a = openS(e)] \lor [OnE(p, e, s) \land$
$\quad \neg(\exists f\ EAt(e, f, s) \land Dst(p, f) \land a = openS(e))]$

### 3.3 Case Representation

When working with first-order specifications of rewards, probabilities, and values, one must be able to assign values to different portions of state space. For this purpose we use

[5]SSAs can often be compiled directly from "effect" axioms that directly specify the effects of actions [15].

the notation $t = case[\phi_1, t_1; \cdots ; \phi_n, t_n]$ as an abbreviation for the logical formula:

$$\bigvee_{i \leq n} \{\phi_i \wedge t = t_i\}.$$

Here the $\phi_i$ are *state formulae* (whose situation term does not use *do*) and the $t_i$ are terms. We sometimes write this $case[\phi_i, t_i]$. Often the $t_i$ will be constants and the $\phi_i$ will partition state space. For example, we may represent our reward function as:

$$case[\forall p, f \; PAt(p,f,s) \supset Dst(p,f) \;,\; 10;$$
$$\exists p, f \; PAt(p,f,s) \wedge \neg Dst(p,f) \;,\; 0]$$

This specifies that we receive a reward of 10 if all people are at their destinations and 0 if not. We introduce the following operators on case statements:

$$case[\phi_i, t_i : i \leq n] \otimes case[\phi_j, v_j : j \leq m] =$$
$$case[\phi_i \wedge \psi_j, t_i \cdot v_j : i \leq n, j \leq m]$$
$$case[\phi_i, t_i : i \leq n] \oplus case[\phi_j, v_j : j \leq m] =$$
$$case[\phi_i \wedge \psi_j, t_i + v_j : i \leq n, j \leq m]$$
$$case[\phi_i, t_i : i \leq n] \ominus case[\phi_j, v_j : j \leq m] =$$
$$case[\phi_i \wedge \psi_j, t_i - v_j : i \leq n, j \leq m]$$

Intuitively, to perform an operation on two case statements, we simply take the cross-product of their partitions and perform the corresponding operation on the resulting paired partitions. Letting each $\phi_i$ and $\psi_j$ denote generic first-order formulae, we can perform the cross-sum $\oplus$ of two case statements in the following manner:

$$case[\phi_1, 10; \; \phi_2, 20] \oplus case[\psi_1, 1; \; \psi_2, 2] =$$
$$case[\phi_1 \wedge \psi_1, 11; \; \phi_1 \wedge \psi_2, 12; \; \phi_2 \wedge \psi_1, 21; \; \phi_2 \wedge \psi_2, 22]$$

Note that the conjunction of two first-order formalae in a partition may be inconsistent, implying that that partition formula can never be satisfied and thus discarded.

We can also represent choice probabilities using case notation. Letting $A(\vec{x})$ be a stochastic action with choices $n_1(\vec{x}), \cdots, n_k(\vec{x})$, we define:

$$pCase(n_j(\vec{x}), A(\vec{x}), s) = case[\phi_1^j(\vec{x}, s), p_1^j; \cdots ; \phi_n^j(\vec{x}, s), p_n^j],$$

where the $\phi_i^j$ partition state space and $p_i^j$ is the probability of choice $n_i(\vec{x})$ being executed under condition $\phi_i^j(\vec{x}, s)$ when the agent executes stochastic action $A(\vec{x})$.

We can likewise use case statements (or sums over case statements) to represent the reward and value functions. We use the notation $rCase(s)$ and $vCase(s)$ to respectively denote these case representations.

### 3.4 Symbolic Dynamic Programming

Suppose we are given a value function in the form $vCase(s)$. Backing up this value function through an action $A(\vec{x})$ yields a case statement containing the logical description of states that would give rise to $vCase(s)$ after doing action $A(\vec{x})$, as well as the values thus obtained. This is analogous to classical goal regression, the key difference being that action $A(\vec{x})$ is stochastic. In MDP terms, the result of such a backup is a case statement representing a Q-function.

There are in fact two types of backups that we can perform. The first, $B^{A(\vec{x})}$, regresses a value function through an action and produces a case statement with *free variables* for the action parameters. The second, $B^A$, existentially quantifies over the free variables $\vec{x}$ in $B^{A(\vec{x})}$ and takes the max over the resulting partitions. The application of this operator results in a case description of the regressed value function indicating the best value that could be achieved by *any* instantiation of $A(\vec{x})$ in the pre-action state.

To define the $B^{A(\vec{x})}$ operator, we first define the *first-order decision theoretic regression (FODTR)* operator [3]:

$FODTR(vCase(s), A(\vec{x})) =$
$\gamma [\oplus_j \{pCase(n_j(\vec{x}), s) \otimes Regr(vCase(do(n_j(\vec{x}), s)))\}]$

(Unlike the original definition of FODTR [3], we do not include the sum of $rCase(s)$ in the FODTR definition since it is useful to keep it separate in this paper.) Thus, we can express the $B^{A(\vec{x})}$ operator as follows:

$$B^{A(\vec{x})}(vCase(s)) = rCase(s) \oplus FODTR(vCase(s), A(\vec{x})) \quad (4)$$

To obtain the result of the $B^A$ operator on this same value function, we need only existentially quantify the free variables $\vec{x}$ in $B^{A(\vec{x})}$, which we denote as $\exists \vec{x} \; B^{A(\vec{x})}$. Since the case notation is disjunctively defined, we can distribute the existential quantification into each individual case partition. However, even if the partitions were mutually exclusive prior to the backup operation, the existential quantification can result in overlap between the partitions. Thus, we modify the existential quantification operator to ensure that the highest possible value is achieved in every state by sorting each partition $\langle \phi(\vec{x}), t_i \rangle$ by the $t_i$ values (such that $i < j \supset t_i > t_j$) and ensuring that a partition can be satisfied only when no higher value partition can be satisfied:

$$\exists \vec{x} \; case[\phi_1(\vec{x}), t_1; \cdots ; \phi_n(\vec{x}), t_n] \equiv$$
$$case[\exists \vec{x} \; \phi_i(\vec{x}) \wedge \bigwedge_{j \leq i} \neg \exists \vec{x} \; \phi_j(\vec{x}), v_i : i \leq n]$$

Assuming that the $rCase(s)$ statement does not depend on the parameters of the action (this can be easily relaxed, if needed) and exploiting the fact $\oplus$ is conjunctively defined, we arrive at the definition of the $B^A$ operator:

$$B^A(vCase(s)) = rCase(s) \oplus \exists \vec{x} \; FODTR(vCase(s), A(\vec{x}))$$

## 4 Approximate Linear Programming for FOMDPs

We now generalize the approximate linear programming techniques from propositional factored MDPs to FOMDPs.

## 4.1 Value Function Representation

We represent the value function as a weighted sum of $k$ *first-order basis functions*, denoted $bCase_i(s)$, each of which is intended to contain *small* partitions of formulae that provide a first-order abstraction of the state space:

$$vCase(s) = \oplus_{i=1}^{k} w_i \cdot bCase_i(s) \qquad (5)$$

As in the propositional case [8, 17], the fact that basis functions are represented compactly using logical formulae to specify partitions of state space means that the value function can be represented very concisely using Eq. 5. Our approach is distinguished by the use of first-order formulae to represent the partition specifications.

We can easily apply our previously defined backup operators $B^{A(\vec{x})}$ and $B^A$ to this representation and obtain some simplification as a result of the structure in Eq. 5. For $B^{A(\vec{x})}$, we substitute value function expression in Eq. 5 into the definition $B^{A(\vec{x})}$ (Eq. 4). Exploiting the fact that $\oplus$ is conjunctively defined and that $Regr(\psi_1 \wedge \psi_2) = Regr(\psi_1) \wedge Regr(\psi_2)$, we push the regression down to the individual basis functions and reorder the summations. As a consequence, we find that the backup of a linear combination of basis functions is simply the linear combination of the FODTR of each basis function:

$$B^{A(\vec{x})}(\oplus_i w_i bCase_i(s)) = \\ rCase(s) \oplus (\oplus_i w_i FODTR(bCase_i(s), A(\vec{x}))) \quad (6)$$

By representing the value function as a linear combination of basis functions, we can often achieve a reasonable approximation of the exact value function by exploiting additive structure inherent in many real-world problems (e.g. additive reward functions or problems with independent subgoals). Unlike exact solution methods where value functions can grow exponentially in size during the solution process and must be logically simplified [3], here we necessarily achieve a compact form of the value function that requires no simplification, just the discovery of good weights.

To apply the $B^A$ operator for a linear combination of basis functions, we let $F_A$ denote the set of indices for basis functions affected by action $A(\vec{x})$ (thus having free variables $\vec{x}$), and let $N_A$ denote the set of indices for basis functions not affected by the action.[6] Then, we again exploit the fact that $\oplus$ is defined by conjunction and push the $\exists \vec{x}$ through to the the sum over case statements with free variables $\vec{x}$. This yields the following definition of $B^A$:

$$B^A(\oplus_i w_i bCase_i(s)) = rCase(s) \oplus (\oplus_{i \in N_A} w_i bCase_i(s)) \\ \oplus \exists \vec{x} \, (\oplus_{i \in F_A} w_i FODTR(bCase_i(s), A(\vec{x})))$$

---

[6]Since an action typically affects only a few fluents and basis functions contain relatively few fluents, only a few of the basis functions will be affected by an action. This fact holds for many domains, including the elevator domain in this paper.

It is important to note that the backup operator results in the "cross sum" $\oplus$ of a collection of case statements, performing the explicit sum only over those case statements affected by the action being backed up. This results in a representation of the backed-up value function that often retains most of its original additive structure.

As a concrete example of the application of backup operators $B^{A(x)}$ and $B^A$, suppose we have a reward function that assigns 10 for being in a state where all people are at their intended destination, and 0 otherwise:[7]

$$rCase(s) = case[\, \forall p, f \ PAt(p,f,s) \supset Dst(p,f), 10 \, ; \, \neg\text{''}, 0\,]$$

Suppose our value function is represented by the following weighted sum of two basis functions (the first basis function indicates whether some person is at a floor that is not their destination, the second whether there is a person on an elevator stopped at a floor that is their destination):

$vCase(s) =$
  $w_1 \cdot case[\, \exists p, f \ PAt(p,f,s) \wedge \neg Dst(p,f), 1 \, ; \, \neg\text{''}, 0\,] \oplus$
  $w_2 \cdot case[\, \exists p, f, e \ Dst(p,f) \wedge OnE(p,f,s) \wedge EAt(e,f,s), 1 \, ;$
  $\neg\text{''}, 0\,]$

Because $down(x)$ is a deterministic action, FODTR is simply the direct regression of each case statement through nature's choice $downS(x)$. Thus, we obtain the following expression for $B^{down(x)}$:

$B^{down(x)}(vCase(s)) =$
  $case[\, \forall p, f \ PAt(p,f,s) \supset Dst(p,f), 10 \, ; \, \neg\text{''}, 0\,]$
  $\oplus \gamma \, w_1 \cdot case[\, \exists p, f \ PAt(p,f,s) \wedge \neg Dst(p,f), 1 \, ; \, \neg\text{''}, 0\,]$
  $\oplus \gamma \, w_2 \cdot case[\, \exists p, f, e \ Dst(p,f) \wedge OnE(p,f,s) \wedge$
    $((EAt(e,f,s) \wedge e \neq x) \vee (EAt(e, fa(f), s) \wedge e = x)), 1 \, ;$
    $\neg\text{''}, 0\,]$

Applying $B^{down}$ yields:

$B^{down}(vCase(s)) =$
  $\exists x \ B^{down(x)}(vCase(s)) \ =$
  $case[\, \forall p, f \ PAt(p,f,s) \supset Dst(p,f), 10 \, ; \, \neg\text{''}, 0\,]$
  $\oplus \gamma \, w_1 \cdot case[\, \exists p, f \ PAt(p,f,s) \wedge \neg Dst(p,f), 1 \, ; \, \neg\text{''}, 0\,]$
  $\oplus \gamma \, w_2 \cdot case[\, \exists x, p, f, e \ Dst(p,f) \wedge OnE(p,f,s) \wedge$
    $((EAt(e,f,s) \wedge e \neq x) \vee (EAt(e, fa(f), s) \wedge e = x)), 1 \, ;$
    $\neg\text{''} \wedge \exists x \forall p, f, e \ \neg Dst(p,f) \vee \neg OnE(p,f,s) \vee$
    $((\neg EAt(e,f,s) \vee e = x) \wedge$
    $(\neg EAt(e, fa(f), s) \vee e \neq x)), 0\,]$

Note that in the computation of $B^{down}$ above, we push the existential into the only case statement containing free variable $x$. We then make the second partition in this case statement mutually exclusive of the first partition to ensure that the highest value gets assigned to every state. This results in a logical description of states that describes the maximum value that can be obtained by applying *any* instantiation of action *down(e)* and subsequently obtaining value dictated by *vCase(s)*. Thus, as noted in [3], the symbolic dynamic programming approach provides *both state and action space abstraction*.

---

[7]We replace the partition formula with $\neg\text{''}$ when the formula is just the negation of its predecessor in the case statement.

### 4.2 First-order Linear Program Formulation

We now generalize the approximate linear programming approach for propositional factored MDPs (i.e., Eq. 3) to first-order MDPs. If we simply substitute appropriate notation, we arrive at the following formulation of the first-order approximate linear programming approach (FOALP):

$$\text{Variables: } w_i \, ; \; \forall i \leq k$$
$$\text{Minimize: } \sum_s \sum_{i=1}^{k} w_i \cdot bCase_i(s)$$
$$\text{Subject to: } 0 \geq B^A(\oplus_{i=1}^{k} w_i \cdot bCase_i(s)) \ominus$$
$$(\oplus_{i=1}^{k} w_i \cdot bCase_i(s)) \, ; \; \forall A, s \quad (7)$$

Unfortunately, while the objective and constraints in the approximate LP for a propositional factored MDP range over a finite number of states $s$, this direct generalization to the FOALP approach for FOMDPs requires dealing with infinitely (or indefinitely) many situations $s$.

Since we are summing over infinitely many situations in the FOALP objective, it is ill-defined. Thus, we redefine the FOALP objective in a manner which preserves the intention of the original approximate linear programming solution for MDPs. In the propositional LP (Eq. 3), the objective weights each state equally and minimizes the sum of the values of all states. Consequently, rather than count ground states in our FOALP objective—of which there would be an infinite number per partition—we suppose that each basis function partition is chosen because it represented a potentially useful partitioning of state space, and thus weight each case *partition* equally. Consequently, we write the above FOALP objective as the following:

$$\sum_s \sum_{i=1}^{k} w_i \cdot bCase_i(s) = \sum_{i=1}^{k} w_i \sum_s bCase_i(s) \quad (8)$$
$$\sim \sum_{i=1}^{k} w_i \sum_{\langle \phi_j, t_j \rangle \in bCase_i} \frac{t_j}{|bCase_i|}$$

With the issue of the infinite objective resolved, this leaves us with one final problem—the infinite number of constraints (i.e., one for every situation $s$). Fortunately, we can work around this since case statements are finite. Since the value $t_i$ for each case partition $\langle \phi_i(s), t_i \rangle$ is constant over all situations satisfying the $\phi_i(s)$, we can explicitly sum over the $case_i(s)$ statements in each constraint to yield a

---

**Input:** A set $c$ of *case* statements being summed and an ordering $R_1 \ldots R_n$ of *all* relations in $c$.
**Output:** The maximum value achievable and the consistent partition which achieves that value.

1. Convert the first-order formulae in each case partition to a set of CNF clauses.

2. For each relation $R \in R_1 \ldots R_n$ (in order):
   (a) Remove all *case* statements in $c$ containing $R$ and store their explicit "cross-sum" $\oplus$ in $tmp$.
   (b) Do the following for each partition in $tmp$:
      - Resolve all clauses on relation $R$.
      - Remove all clauses containing $R$ (sound because we are doing ordered resolution).
      - If a resolvent of $\emptyset$ exists in this partition, remove this partition from $tmp$ and continue.
      - Remove dominated partitions whose clauses are a superset of another partition with greater value.
   (c) Insert $tmp$ back into $c$.

3. Find the maximum partition in each of the case statements in $c$ and return it along with its corresponding value.

Figure 1: The first-order factored max (FOMAX) algorithm for finding the max consistent partition in a first-order cost network.

---

single case statement. For this "flattened" case statement, we can easily verify that the constraint holds in each of the finite number of resulting first-order partitions of state space.

Furthermore, we can avoid generating the full set of case partitions in the explicit sum by using constraint generation [17]. But before we formally define the algorithm for solving first-order LPs via constraint generation, we first outline an important subroutine.

**First-order Cost Network Maximization**

One of the most important computations in the approximate FOMDP solution algorithms is efficiently finding the consistent set of case partitions which maximizes a first-order cost network of the following form:

$$\max_s (\oplus_j \, case_j(s)) \quad (9)$$

Recall that this is precisely the form of the constraints (the backed up linear value function) in Equation 7. The FOMAX algorithm given in Figure 1 efficiently carries out this computation. It is similar to variable elimination [4], except that we use first-order ordered resolution in place of propositional ordered resolution.[8]

---

[8]We note that the ordered resolution strategy is not refutation-complete as it may loop infinitely at an intermediate relation elimination step before finding a latter relation with which to resolve a contradiction. To prevent infinite looping, we simply bound the number of inferences performed at each relation elimination step. While a missed refutation in the FOMAX algorithm may result in the generation of unneccessary constraints during the FOCG algorithm, and this in turn may overconstrain the set of feasible

**Input:** An LP objective $obj$, a convergence tolerance $tol$, and a set of first-order constraints, each encoded as $0 \geq \max_s \oplus_i case_i(s)$ where there are $w_i$ variables embedded in the values of the case partitions.
**Output:** Weights for the optimal LP solution (within tolerance $tol$).

1. Initialize the weights: $w_i = 0$ ; $\forall i \leq k$
2. Initialize the LP constraint set: $C = \emptyset$
3. Initialize $C_{new} = \emptyset$
4. For each constraint inequality:
    (a) Calculate $\varphi = \arg \max_s \oplus_i case_i(s)$ using FO-MAX.
    (b) If $eval(\varphi) \geq tol$, let $c$ encode the linear constraint for $0 \geq \oplus_i case_i(\varphi)$.
    (c) $C_{new} = C_{new} \cup \{c\}$
5. If $C_{new} = \emptyset$, terminate with $w_i$ as the solution to this LP.
6. $C = C \cup C_{new}$
7. Solve the LP with objective $obj$ and updated constraints $C$.
8. Return to step 3.

Figure 2: The first-order constraint generation (FOCG) algorithm for solving a first-order linear program.

**First-Order Constraint Generation**

In the first-order constraint generation (FOCG) algorithm given in Figure 2, we initialize our LP with an initial setting of weights, but no constraints. Then we alternate between generating constraints based on maximal constraint violations at the current solution and re-solving the LP with these additional constraints. This process repeats until no constraints are violated and we have found the optimal solution. In practice, this approach typically generates *far* fewer constraints than the full set. And using FOCG, we can now efficiently solve the FOALP given in Equation 7.

To make the FOMAX and FOCG algorithms more concrete, we provide a simple example of finding the most violated constraint in Figure 3.

### 4.3 Obtaining Error Bounds

Following [17], we can also compute an upper bound on the error of the approximated value function:

$$\max_s vCase^*(s) \ominus vCase_{\pi_{greedy}(\oplus_i w_i \cdot bCase_i)}(s) \leq$$

$$\frac{\gamma}{1-\gamma} \min_A \max_s \left[ (\oplus_i w_i \cdot bCase_i(s)) \ominus B^A (\oplus_i w_i \cdot bCase_i(s)) \right] \quad (10)$$

The inner $\max_s$ can be efficiently computed by the FO-MAX algorithm (Figure 1) leaving only a few simple arithmetic operations to carry out. This yields an efficient means of computing a strict upper bound on the error of executing a greedy policy according to an approximated value function that applies *equally to all domain instantiations*.

## 5 Experimental Results

We have applied the first-order approximate linear programming algorithm to the FOMDP domain of elevator scheduling. We refer to Section 3 for a rough description of the FOMDP specification of certain aspects of this domain. To this we have added a rich set of features used in elevator planning domains [12]: $VIP(p)$ indicating that person $p$ is a VIP, $Group(p,g)$ indicating that a person $p$ is in group $g$ (persons in certain differing groups should not share the same elevator), and $Attended(p)$ to indicate that person $p$ requires someone to accompany them in the elevator. We have also added action costs of 1 for *open*, *up*, *down* and a simple *noop* action with cost 0.[9] We require that all policies observe typical elevator scheduling constraints [12] by specifying appropriate action preconditions; for example, the elevator cannot reverse direction if it has a passenger whose destination is in the current direction of travel.

Given this domain specification, we have defined a multicriteria reward function $rCase(s)$ with positive additive rewards for satisfying each of the following conditions:

$+2 : \forall p, f\ PAt(p, f, s) \supset Dst(p, f)$
$+2 : \forall p, f\ Dst(p, f) \wedge \neg PAt(p, f, s) \supset \exists e\ On(p, e, s)$
$+2 : \forall p, f\ VIP(p) \wedge PAt(p, f, s) \supset Dst(p, f)$
$+2 : \forall p, f\ VIP(p) \wedge Dst(p, f) \wedge \neg PAt(p, f, s)$
$\qquad \supset \exists e\ OnE(p, e, s)$
$+4 : \forall p, e\ OnE(p, e, s) \wedge Attended(p) \supset \exists p_2 OnE(p_2, e, s)$
$+8 : \forall p_1, p_2, g_1, g_2, e\ OnE(p_1, e, s) \wedge OnE(p_2, e, s)$
$\qquad \wedge p_1 \neq p_2 \wedge Group(p_1, g_1) \wedge Group(p_2, g_2) \supset g_1 = g_2$

In our FOALP solution, we have used a set of basis functions that represent each of the additive reward criteria above, in addition to one constant basis function. We experimented with a variety of more complex compound basis functions in addition to the ones above but found that they received negligible weight and did not affect the resulting policy. Thus, it seems that a linear combination of these six basis functions may approximate a reasonable amount of structure in the optimal value function.

We evaluated two simple myopic policies that take the best action from the current state based on one- and two-step lookahead. We also evaluated several intuitive, heuristic policies that seem to offer reasonable performance in the elevator domain. The basic heuristic policy always picks up passengers at each floor. The other policies are specified by adding the following constraints to this basic policy in various combinations: always prioritize picking up VIPs

---

solutions, this has not led to infeasibility problems in practice.

[9]Since these costs are constant for each action, the backup operators can be easily modified to handle action costs by subtracting the constant cost from *one* of the cases being backed up.

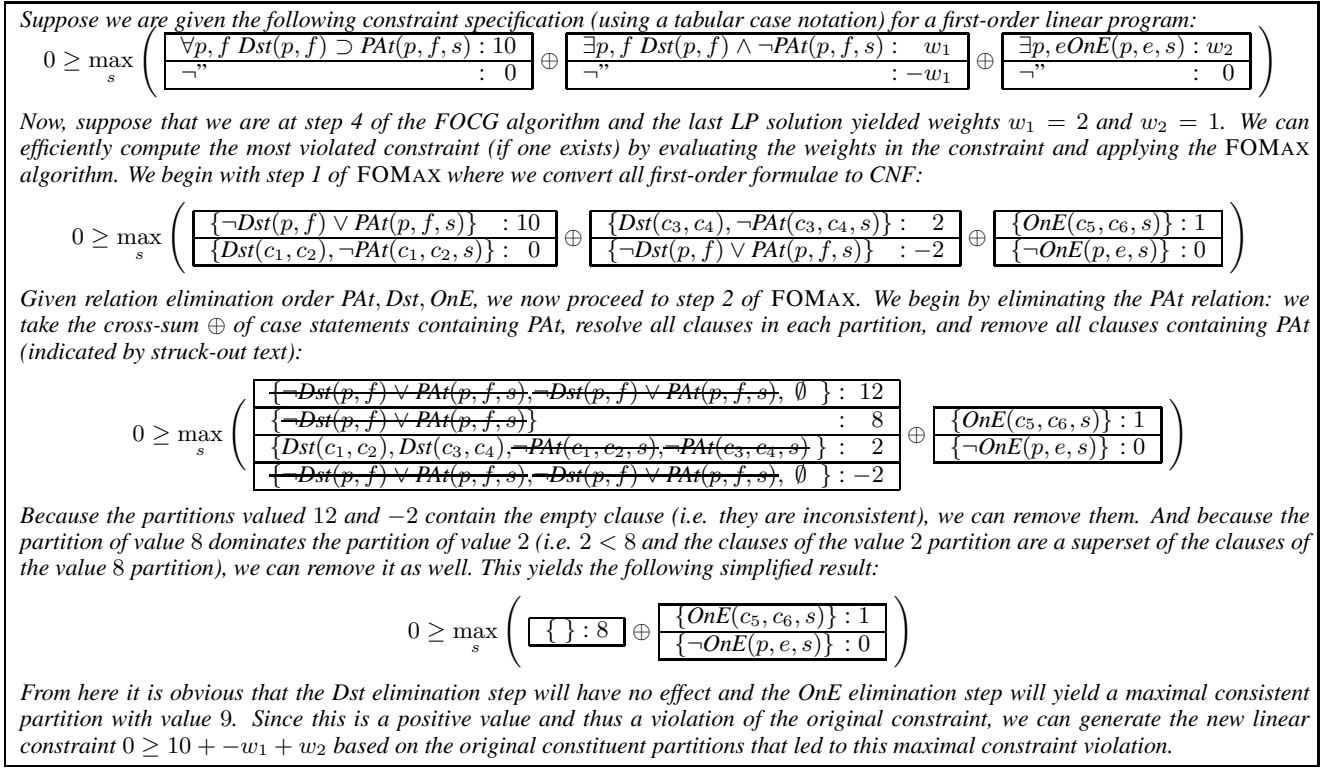

Figure 3: An example of using FOMAX to find the maximally violated constraint during an iteration of the FOCG algorithm.

over regular passengers (V); never allow an attended person to be unaccompanied in the elevator (A); and never allow a group conflict in the elevator (G).

We used the Vampire [16] theorem prover and the CPLEX LP solver in our FOALP implementation.[10] The performance of various policies, measured using the average total discounted reward obtained over 100 simulated trials of 50 stages each, are shown in Table 1. While the results have relatively high standard deviation due to the stochasticity of the domain, the difference in performance between FOALP with six basis functions and the best of the heuristic policies is statistically significant at the 93% confidence level for each domain size. Thus, we can conclude that the FOALP approach with six basis functions outperforms the heuristic policies with some confidence. This appears to be due to the fact that the additive value function allows FOALP to tradeoff various domain features while the hard constraints must be applied in all situations (leading to suboptimal performance in many cases) and the myopic policies cannot see far enough ahead to make good choices for the long-term. We note that while the upper bounds on error are not tight, there is nevertheless a strong correlation between these values and actual policy performance.

| Policy | F5 | F10 | F15 | ME |
|---|---|---|---|---|
| { }, {A} | $116 \pm 28$ | $106 \pm 27$ | $105 \pm 28$ | N/A |
| {V}, {V,A} | $115 \pm 30$ | $108 \pm 30$ | $107 \pm 28$ | N/A |
| {G}, {A,G} | **$125 \pm 24$** | **$119 \pm 21$** | $114 \pm 20$ | N/A |
| {V,G}, {V,A,G} | $119 \pm 30$ | $114 \pm 24$ | **$115 \pm 23$** | N/A |
| Myopic 1 | $118 \pm 10$ | $119 \pm 9$ | $120 \pm 13$ | N/A |
| Myopic 2 | **$123 \pm 12$** | **$122 \pm 5$** | **$120 \pm 12$** | N/A |
| FOALP {1,2} | $133 \pm 31$ | $114 \pm 32$ | $112 \pm 23$ | 177 |
| FOALP {3,4} | $148 \pm 26$ | $129 \pm 23$ | $117 \pm 23$ | 159 |
| FOALP {5} | $147 \pm 26$ | $126 \pm 17$ | $120 \pm 17$ | 146 |
| FOALP {6} | **$154 \pm 25$** | **$130 \pm 19$** | **$125 \pm 19$** | 92 |

Table 1: The performance of various policies in elevator domains with 5, 10, and 15 floors (denoted F5, F10, and F15, respectively), with passenger arrivals distributed according to $N(0.1, 0.35)$ per time-step, and discount rate $\gamma = 0.9$. Heuristic policies, shown at the top, are designated by the sets of constraints used: (V) prioritize VIP, (A) no attended conflict, and (G) no group conflict. For example, the empty set policy {} uses no constraints and picks up everyone at each floor it passes. The one- and two-step lookahead myopic policies are labeled *Myopic 1* and *2*. FOALP policies are labeled using the number of basis functions used, counting from the top of the above list of reward criteria (e.g., *FOALP {3,4}* represents two FOALP-derived policies using the first three and first four basis functions, respectively). Policies are grouped when they turn out to be identical (e.g., FOALP-3 and FOALP-4 are identical, as are heuristic policies {$V$} and {$V, A$}). Performance is measured using the the sum of discounted rewards obtained by time step 50, averaged over 100 random simulations. The best results for each group of policies and number of floors are in boldface. FOALP error bounds are computed according to Equation 10 and are shown in the max error (ME) column.

---

[10]We note that while the full FOALP specification with six basis functions contained over 250,000 constraints, only 40 of these constraints needed to be generated to determine the optimal solution. Nonetheless, a considerable time is spent in theorem proving deriving and ensuring the consistency of constraints. We note that running times ranged from 5 minutes for one basis function to 120 minutes for six basis functions on a 2Ghz Pentium-IV machine.

## 6 Related Work

While our work incorporates elements of symbolic dynamic programming (SDP) [3], it uses this in an efficient approximation framework that represents the value function compactly and avoids the need for logical simplification. This is in contrast to an exact value iteration framework that proves intractable in many cases for SDP, FOVIA [10], and ReBel [11] due to the combinatorial explosion of the value function representation and the need to perform complex logical simplifications. FOALP uses a model-based approach to avoid domain instantiation unlike the non-SDP approaches [5, 7] which need to instantiate domains. In addition, FOALP does not use inductive methods [5, 6] that generally require substantial simulation to generalize from experience. The work by Guestrin *et al.* [7] is the only other RMDP approximation work that can provide bounds on policy quality, but these are PAC-bounds under the assumption that the probability of domains falls off exponentially with their size. In contrast, the policy bounds obtained by FOALP apply equally to all domains.

## 7 Concluding Remarks

We have introduced a novel approximate linear programming technique for FOMDPs that represents an approximately optimal value function as a linear combination of first-order basis functions. This technique allows us to efficiently solve for an approximation of the value function using off-the-shelf theorem prover and LP software. It offers us the advantage that we can approximately solve FOMDPs independent of a domain instantiation and it additionally allows us to determine upper bounds on approximation error that apply equally to all possible domain instantiations. We have applied our algorithm to the task of elevator scheduling and have found that it is not only efficient, but that its policies outperform a number of intuitive, heuristically-guided policies for this domain.

An interesting question for future work is whether the uniform weighting of partitions in the FOALP objective is the best approach. It would be informative to explore alternative FOALP objective specifications and to evaluate their impact on value function quality over a variety of domains. In addition, it is an interesting question as to whether dynamic reweighting schemes could improve solution quality by identifying state partitions with large error and adjusting their weights so they receive more emphasis on the next LP iteration. Altogether, such improvements to FOALP could bolster an already promising approach for efficiently and compactly approximating FOMDP value functions.